\documentclass{article}
\usepackage{spconf,amsmath,graphicx}
\usepackage{multicol,multirow}

\usepackage{booktabs}  
\usepackage{siunitx}   
\usepackage{multirow}  

\usepackage{enumitem}
\setlist{nosep, leftmargin=14pt}

\usepackage{mwe} 

\usepackage{makecell}
\usepackage{amssymb}

\title{MIQ-SAM3D: From Single-Point Prompt to Multi-Instance Segmentation via Competitive Query Refinement}
\twoauthors
 {Jierui Qu}
	{National University of Singapore\\
	College of Design and Engineering\\
	117575, Singapore}
 {Jianchun Zhao\sthanks{Corresponding Author, email: zhao\_jianchun@stu.xjtu.edu.cn.}}
	{Xi'an Jiaotong University\\
	School of Electrical Engineering\\
	Xi'an 710049, China}
\begin{document}
\ninept
\maketitle
\begin{abstract}
Accurate segmentation of medical images is fundamental to tumor diagnosis and treatment planning. SAM-based interactive segmentation has gained attention for its strong generalization, but most methods follow a single-point-to-single-object paradigm, which limits multi-lesion segmentation. Moreover, ViT backbones capture global context but often miss high-fidelity local details. We propose MIQ-SAM3D, a multi-instance 3D segmentation framework with a competitive query optimization strategy that shifts from single-point-to-single-mask to single-point-to-multi-instance. A prompt-conditioned instance-query generator transforms a single point prompt into multiple specialized queries, enabling retrieval of all semantically similar lesions across the 3D volume from a single exemplar. A hybrid CNN-Transformer encoder injects CNN-derived boundary saliency into ViT self-attention via spatial gating. A competitively optimized query decoder then enables end-to-end, parallel, multi-instance prediction through inter-query competition. On LiTS17 and KiTS21 dataset, MIQ-SAM3D achieved comparable levels and exhibits strong robustness to prompts, providing a practical solution for efficient annotation of clinically relevant multi-lesion cases.
\end{abstract}
\section{Introduction}
Accurate medical image segmentation is foundational to modern clinical practice. It underpins disease diagnosis, treatment planning, and prognostic assessment~\cite{antonelli2022medical}. In recent years, deep learning—particularly convolutional neural networks (CNNs) such as U‑Net and its variants—has achieved strong performance in three‑dimensional (3D) medical image segmentation~\cite{cicek20163da,ronneberger2015uneta}. However, most of these models remain task‑specific and generalize poorly. Moreover, the inherently local receptive fields of CNNs limit their ability to model global context and long‑range dependencies~\cite{manzari2024befunet,shaker2024unetra,park2025esunet,huang2025revisiting}.

To overcome the limitations of traditional CNNs in understanding global context, the research community has turned to exploring Vision Transformer (ViT) architectures with robust long-range dependency modeling capabilities~\cite{azad2023foundational}. Among these, the large-scale pre-trained foundation model based on ViT—the Segment Anything Model (SAM)—has garnered significant attention for its exceptional zero-shot and promptable segmentation capabilities on natural images~\cite{kirillov2023segment,zhang2025hierarchical,wu2025samaware}. This breakthrough has rapidly spurred extensive research aimed at adapting SAM's powerful capabilities to 3D medical image segmentation through strategies like parameter-efficient fine-tuning (PEFT), bridging the significant domain gap between natural and medical images~\cite{mazurowski2023segment}. However, current research is constrained by two core limitations: most frameworks adhere to a “single-point prompt, single-object segmentation” paradigm, struggling to address clinically prevalent diffuse lesion instance segmentation tasks; moreover, their inherent ViT backbone networks sacrifice high-fidelity local details crucial for ambiguous boundaries while capturing global context.

\begin{figure*}[htbp]
\centering
\includegraphics[width=0.99\textwidth]{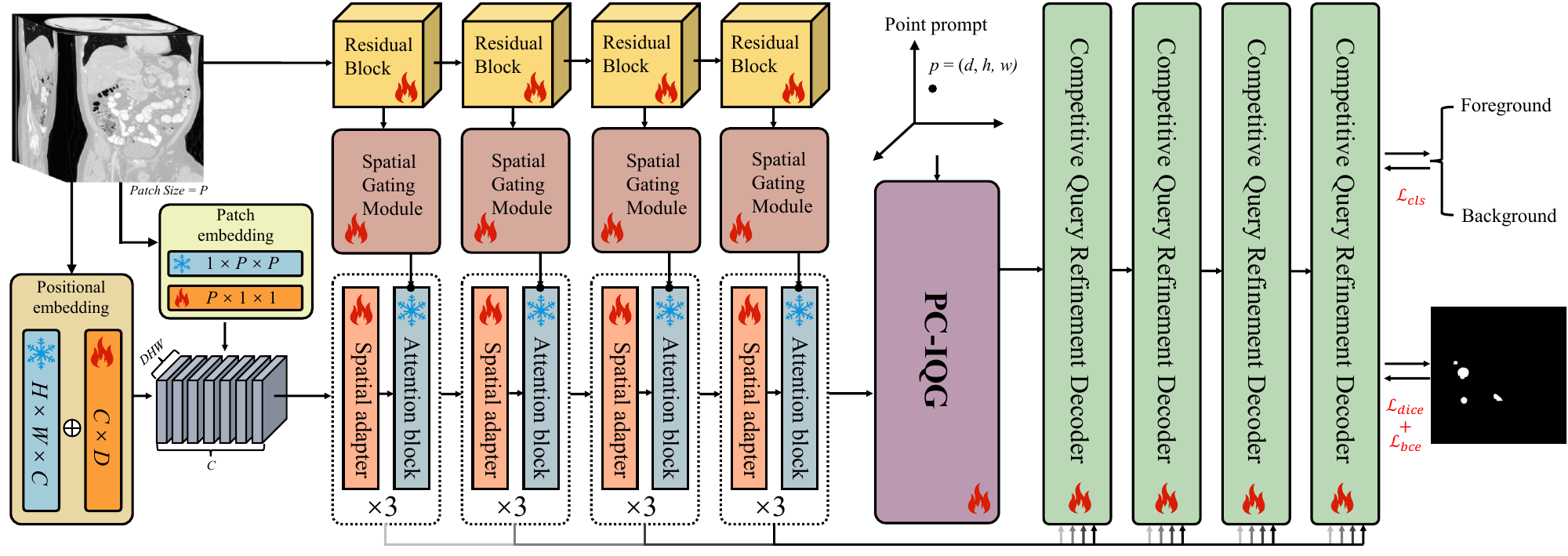}
\caption{Overview of our proposed method for MIQ-SAM3D.This framework consists of a hybrid CNN-Transformer encoder, a prompt-conditioned instance query generator (PC-IQG), a competitive query-optimized decoder (CQRD), and an instance prediction head.} \label{fig:overview}
\end{figure*}

To systematically address the aforementioned limitations, this paper proposes the MIQ-SAM3D segmentation framework, which can simultaneously handle multi-instance objects and perceive high-fidelity details. We introduce a prompt-initiated instance segmentation paradigm inspired by modern detection models~\cite{cheng2022maskedattention}, which transforms a single-point user prompt into a set of dynamic object queries via a prompt-conditioned instance query generator (PC-IQG). These queries are then fed into a Competitive Query Refinement Decoder, enabling the identification of all similar instances throughout the 3D volume using the user-clicked lesion as an “example.” Additionally, we designed a hybrid CNN-Transformer encoder. A parallel 3D CNN branch specializes in capturing local features, while a Spatial Gate Module deeply integrates high-fidelity details into the self-attention computation of the ViT backbone, significantly improving segmentation accuracy for blurred boundaries.

The main contributions of this paper can be summarized as follows:

\begin{enumerate}
    \item We propose a prompt-initiated MIQ-SAM3D segmentation framework that achieves comparable performance in segmenting multiple disjoint object instances with a single point prompt. This is accomplished through our designed PC-IQG and competitive query optimization decoder.
    \item We designed an efficient hybrid CNN-Transformer encoder whose core gated cross-fusion mechanism synergistically leverages CNN's local advantages and ViT's global perspective, achieving outstanding segmentation performance with high-fidelity details.
    \item Our approach was validated on multiple public 3D tumor segmentation benchmarks. Experimental results demonstrate comparable performance on multi-instance segmentation tasks.
\end{enumerate}

\section{Method}
The overall architecture of the multi-instance 3D segmentation framework proposed in this study is shown in Fig.~\ref{fig:overview}. For the input 3D medical image $\mathbf{X} \in \mathbb{R}^{D \times H \times W}$, local feature extraction and global feature modeling are performed via a dual-branch CNN-Transformer encoder. User prompts $\mathbf{p}$ are fed into the Prompt-Conditioned Instance Query Generato(PC-IQG) module to extract seed prototypes and dynamically generate $N$ instance queries ${Q}_{inst}$. ${Q}_{inst}$ and image features undergo iterative optimization via Competitive Query Refinement Decode(CQRD), enabling competing queries to independently lock onto distinct target instances. Finally, the optimized queries are fed into two parallel prediction heads, enabling end-to-end prediction from single-point prompting to multi-instance segmentation.

\subsection{Hybrid CNN-Transformer Encoder}

\begin{figure}[t]
\centering
\includegraphics[width=0.99\linewidth]{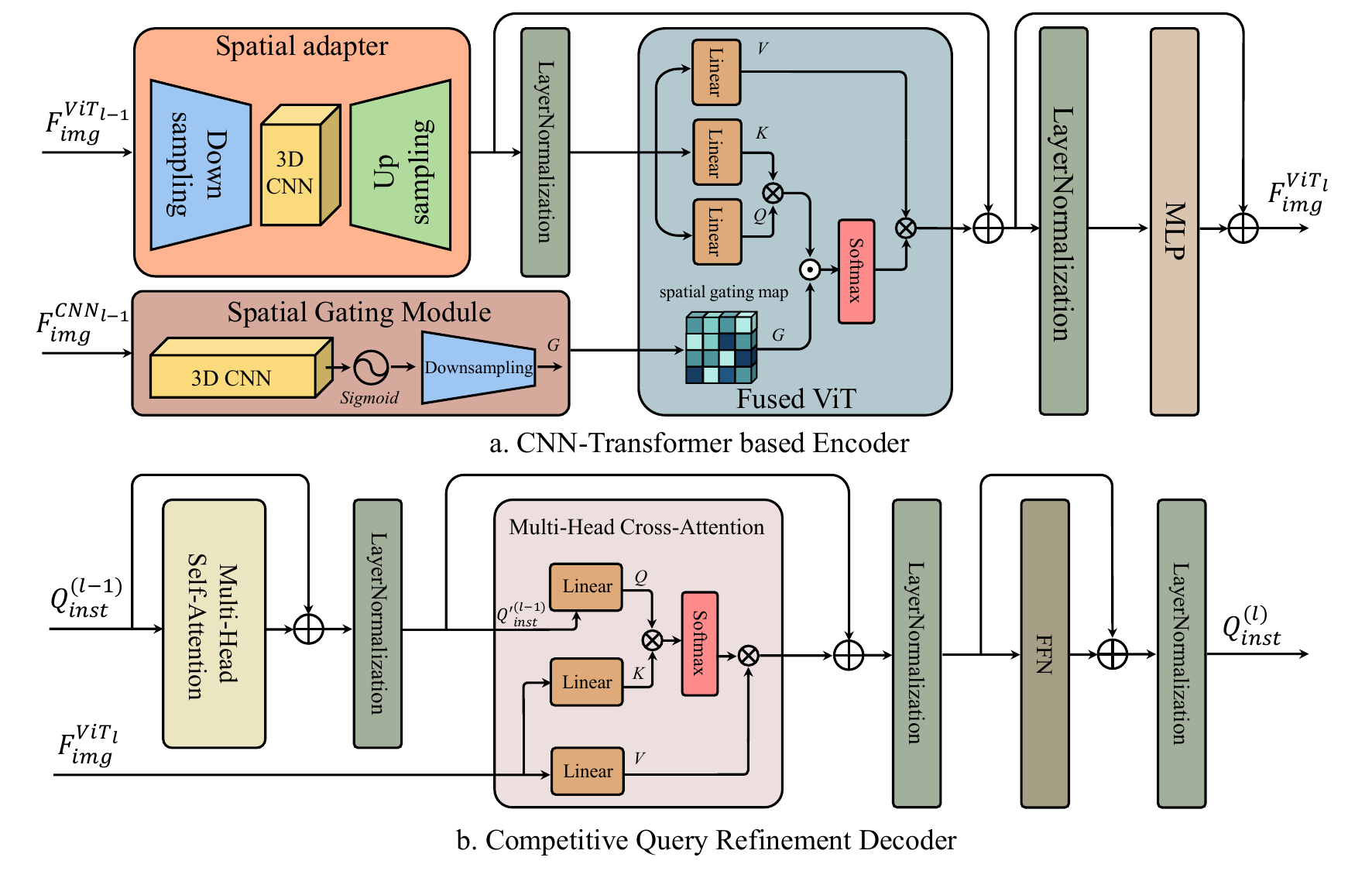}
\caption{Structures of the encoder and decoder. (a)Structure diagram of the dual-branch hybrid CNN-Transformer encoder, which uses a spatial gating module to fuse dual-channel features. (b)Structure diagram of CQRD.} \label{fig:encoder-decoder}
\end{figure}

The hybrid CNN-Transformer encoder adopts a dual-branch architecture. Through the Spatial Gating Module, it integrates the local detail features extracted by the Residual Block into the ViT module, guiding the extraction of global features, as shown in Fig.~\ref{fig:encoder-decoder}(a). The ViT branch inherits the pre-trained weights from the 3DSAM-adapter~\cite{gong20233dsamadaptera}, responsible for capturing long-range dependencies and global semantic information. Most of its parameters remain frozen to preserve pre-trained knowledge. The CNN branch employs lightweight residual blocks trained from scratch to extract high-frequency details like edges and textures across multiple scales. These features are fed into the Spatial Gating Module to generate the spatial gating map $G$:

\begin{equation}
    G=Sigmoid(Conv_{1\times1\times1}(F_{img}^{CNN_{l-1}}))
\end{equation}

In the Fused ViT block, input features undergo linear transformations to yield queries Q, keys K, and values V, and compute the attention score matrix: $A = \frac{QK^T}{\sqrt{d_k}}$. The gating graph $G$ is downsampled to a dimension compatible with the attention score matrix and applied to the attention scores:

\begin{equation}
    Attention_{fued}=\left(Softmax\left(A \odot Downsample\left(G\right)\right)\right)V
\end{equation}

The symbol $\odot$ denotes element-wise multiplication. This gated fusion mechanism generates feature representations for subsequent instance retrieval that combine global context with high-fidelity details.

\subsection{Prompt-Conditioned Instance Query Generator(PC-IQG)}
This paper proposes the PC-IQG module, which uses the features of a user-specified single point location as a category prototype. Based on this, it generates multiple specialized instance detectors to retrieve and segment all semantically similar objects throughout the entire 3D volume. The user-provided 3D coordinates $p = (d, h, w)$ of the query point are input into the PC-IQG module alongside features from the hybrid encoder. A visual sampler~\cite{gong20233dsamadaptera} extracts a seed prototype vector $v_{seed} \in R^C$ from the feature map at this location.

The seed prototype $v_{seed}$ encodes the semantic category, morphological features, and local contextual information of the target instance clicked by the user, serving as the key conditional signal for subsequent query generation. Subsequently, $v_{seed}$ is fed into a multi-layer perceptron-based query generation network $G_\theta$, which expands the single seed prototype into N diverse instance queries $Q_{inst} \in R^{N \times C}$ through nonlinear transformations:

\begin{equation}
    Q_{inst}=\mathcal{G}_\theta(v_{seed})
\end{equation}

The dynamic adaptation mechanism endows the model with strong zero-shot generalization capabilities, enabling it to perform segmentation tasks without retraining for new categories.

\subsection{Competitive Query Refinement Decoder(CQRD)}
CQRD employs a multi-layer Transformer decoder architecture that progressively optimizes instance representations through competitive-collaborative mechanisms among queries and interactions with image features, as illustrated in Fig.~\ref{fig:encoder-decoder}(b). This decoder consists of four identically structured decoding layers stacked sequentially. At layer l, the input instance queries $Q_{inst}^{\left(l-1\right)}$ undergo an inter-query self-attention module to suppress redundant predictions for the same instance, ensuring different queries lock onto distinct instance targets, yielding ${Q^{\prime}}_{inst}^{\left(l-1\right)}$. Subsequently, ${Q^{\prime}}_{inst}^{\left(l-1\right)}$ interacts with the image features output by the encoder through a multi-head cross-attention module. Each query selectively aggregates spatial information relevant to its assigned instance from the global feature map, thereby progressively refining the instance's positional and morphological representations:

\begin{equation}
    {Q^{\prime\prime}}_{inst}^{\left(l-1\right)}=\mathrm{CrossAttention}\left({Q^{\prime}}_{inst}^{\left(l-1\right)},F_{img}^{ViT_l}\right)+{Q^{\prime}}_{inst}^{\left(l-1\right)}
\end{equation}

Finally, the query's expressive power is further enhanced through nonlinear transformations in the feedforward network, producing the final query representation of this layer:

\begin{equation}
    Q_{inst}^{\left(l\right)}=\mathrm{FFN}\left({Q^{\prime\prime}}_{inst}^{\left(l-1\right)}\right)+{Q^{\prime\prime}}_{inst}^{\left(l-1\right)}
\end{equation}

After $L$ layers of iterative optimization, the final query $Q_{inst}^L$ is fed into two parallel prediction heads to generate segmentation results. The classification head predicts the probability distribution of categories corresponding to each query, distinguishing foreground instances from background. The mask head generates instance masks by performing dot product operations between query embeddings and image features.

\begin{figure*}[t]
\centering
\includegraphics[width=0.99\linewidth]{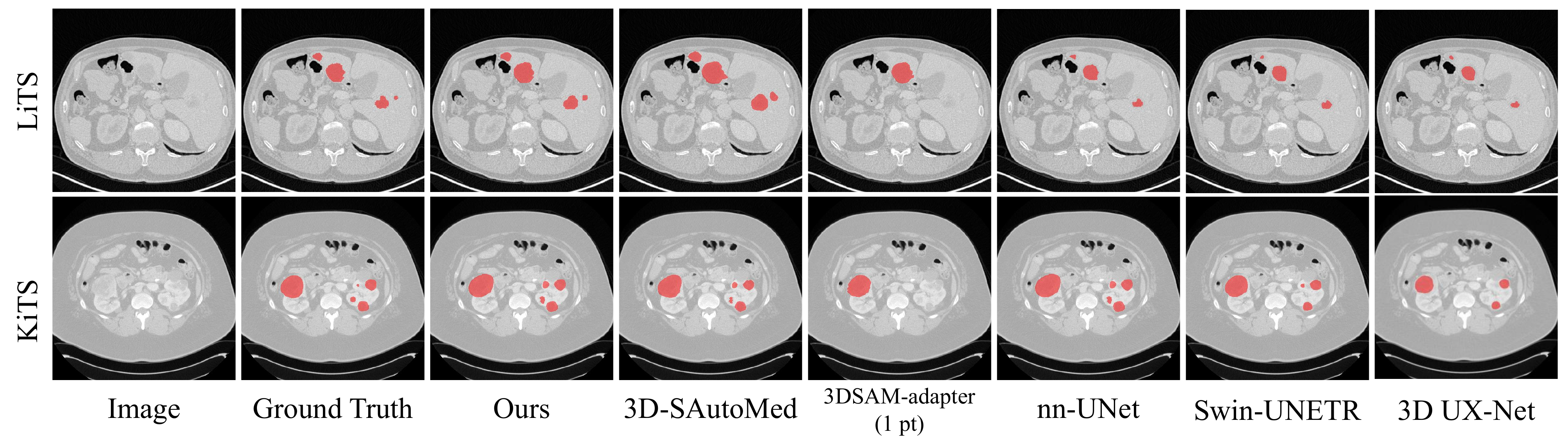}
\caption{Comparison with classical medical image segmentation methods on the LiTS17 and KiTS21 datasets. The evaluation metrics for dice scores and normalized surface dice (NSD) were reported.} \label{fig:visual}
\end{figure*}

\subsection{Loss Function}
Since the number of instances $M$ in the ground truth annotations is typically much smaller than $N$ and instances lack a fixed order, this paper employs a set prediction loss based on bipartite graph matching~\cite{carion2020endtoenda} for end-to-end training.Let the set of true labels be $y=y_1,y_2,\ldots,y_M$, where each $y_j = (c_j, m_j)$ contains the category label $c_j$ and the 3D mask $m_j$. Let the prediction set of the model be $\hat{y}=\hat{y}_1,\hat{y}_2,\ldots,\hat{y}_M$, where each $\hat{y}_j = (\hat{c}_j, \hat{m}_j)$. Define the matching cost between prediction i and actual instance j as $\mathcal{C}\left(i,j\right)$:

\begin{align}
    \mathcal{C}\left(i,j\right)=-\lambda_{cls}\cdot P_i\left(c_j\right)&+\lambda_{dice}\cdot \mathcal{L}_{dice}\left(\hat{m}_i,m_j\right)\notag\\
    &+\lambda_{bce} \cdot \mathcal{L}_{bce}\left(\hat{m}_i,m_j\right)
\end{align}

Here, $P_i(c_j)$ denotes the probability of predicting that i belongs to category $c_j$. The Hungarian algorithm is used to find the permutation that minimizes the total matching cost.

\begin{equation}
    \hat{\sigma} = \arg\min_{\sigma}\; \sum_{i=1}^{N} \mathcal{C}\!\left(\hat{y}_{\sigma(i)},\, y_i\right)
\end{equation}

Thus establishing an optimal bijection between predictions and ground truth, the total loss function is computed only for successfully matched prediction pairs:

\begin{align}
    \mathcal{L}_{total}=\sum_{i=1}^{N}[\lambda_{cls}\mathcal{L}_{cls}&\left(\hat{p}_{\hat{\sigma}\left(i\right)},c_i\right)+\lambda_{dice}\mathcal{L}_{dice}\left(\hat{m}_{\hat{\sigma}\left(i\right)},m_i\right) \notag\\
    +&\lambda_{bce}\mathcal{L}_{bce}\left(\hat{m}_{\hat{\sigma}\left(i\right)},m_i\right)]
\end{align}

\section{Experience}
\subsection{Datasets}
We employed two publicly available datasets for tumor segmentation: the KiTS21 dataset ~\cite{heller2021state} and the LiTS17 dataset ~\cite{bilic2023liver}, both of which feature multiple tumors. The original datasets included organ and tumor segmentation labels, but we exclusively utilized tumor labels for training and testing. The datasets were randomly partitioned into 70\%, 10\%, and 20\% for training, validation, and testing, respectively. The Dice coefficient (Dice) and Normalized Surface Dice (NSD) are used as evaluation metrics.

\subsection{Experimental results and visualization}
To validate the effectiveness of the proposed framework in multi-instance segmentation tasks for 3D medical images, we selected five representative methods for comparison: nn-UNet ~\cite{isensee2021nnunet}, Swin-UNETR~\cite{hatamizadeh2022swin}, 3D UX-Net~\cite{lee20233d}, 3DSAM-adapter(1 pt)~\cite{gong20233dsamadaptera}, and 3D-SAutoMed~\cite{liang20243dsautomeda}. Detailed results are shown in Table~\ref{tab:comparison}. Our method achieved an optimal Dice coefficient of 60.47\% on the Liver Tumor dataset, which was 1.37 percentage points higher than 59.10\% of the traditional SOTA method nn-UNet and 4.66 percentage points higher than 55.81\% of the similar SAM adaptation method 3DSAM-adapter. It is worth noting that liver tumors usually present the characteristics of multiple occurrence and blurred boundaries. This result fully validates the effectiveness of the hybrid CNN-Transformer encoder in capturing fine boundary details and the performance of the prompt conditional query mechanism in multi-instance scenarios. On the Kidney Tumor dataset, our approach attains state-of-the-art performance with a Dice score of 74.83\%. The NSD metric demonstrates comparable levels across both datasets. Fig.~\ref{fig:visual} demonstrates typical multi-instance segmentation visualizations. Under single-point-input conditions, our method achieves segmentation of multiple tumor regions with clear and complete segmentation boundaries.

\begin{table*}
\centering
\caption{Comparison with classical medical image segmentation methods on the LiTS17 and KiTS21 datasets. The evaluation metrics for dice scores and normalized surface dice (NSD) were reported.}
\label{tab:comparison}
\begin{tabular}{ll S[table-format=2.2] S[table-format=2.2] S[table-format=2.2] S[table-format=2.2]}
\toprule
\multirow{2}{*}{Method} & \multirow{2}{*}{Framework} & \multicolumn{2}{c}{LiTS17 (Liver Tumor)} & \multicolumn{2}{c}{KiTS21 (Kidney Tumor)} \\
\cmidrule(lr){3-4} \cmidrule(lr){5-6}
& & {Dice (\%) $\uparrow$} & {NSD (\%) $\uparrow$} & {Dice (\%) $\uparrow$} & {NSD (\%) $\uparrow$} \\
\midrule
nn-UNet~\cite{isensee2021nnunet} & U\,-net & 59.10 & \textbf{77.35} & 73.79 & 81.24 \\
Swin-UNETR~\cite{hatamizadeh2022swin} & U\,-net & 52.10 & 68.07 & 66.04 & 74.29 \\
3D-UX-Net~\cite{lee20233d} & U\,-net & 48.27 & 65.72 & 60.49 & 68.23 \\
3DSAM-adapter (1 pt)~\cite{gong20233dsamadaptera} & SAM & 55.81 & 70.19 & 73.08 & \textbf{85.73} \\
3D-SAutoMed~\cite{liang20243dsautomeda} & SAM & 58.92 & 74.83 & 74.27 & 82.09 \\
Ours & SAM & \textbf{60.47} & 74.61 & \textbf{74.83} & 79.57 \\
\bottomrule
\end{tabular}
\end{table*}

\subsection{Ablation Studies}
\begin{table}
\centering
\caption{Ablation study on LiTS17 dataset.}
\label{tab:ablation}
\begin{tabular}{l S[table-format=2.2] S[table-format=2.2]}
\toprule
Method & {Dice (\%) $\uparrow$} & {NSD (\%) $\uparrow$} \\
\midrule
Ours (Full Model) & 60.47 & 74.61 \\
Ours w/o PC-IQG + CQRD & 56.72 & 75.85 \\
Ours w/o CNN branches & 58.95 & 68.80 \\
\bottomrule
\end{tabular}
\end{table}

\begin{figure}[t]
\centering
\includegraphics[width=0.99\linewidth]{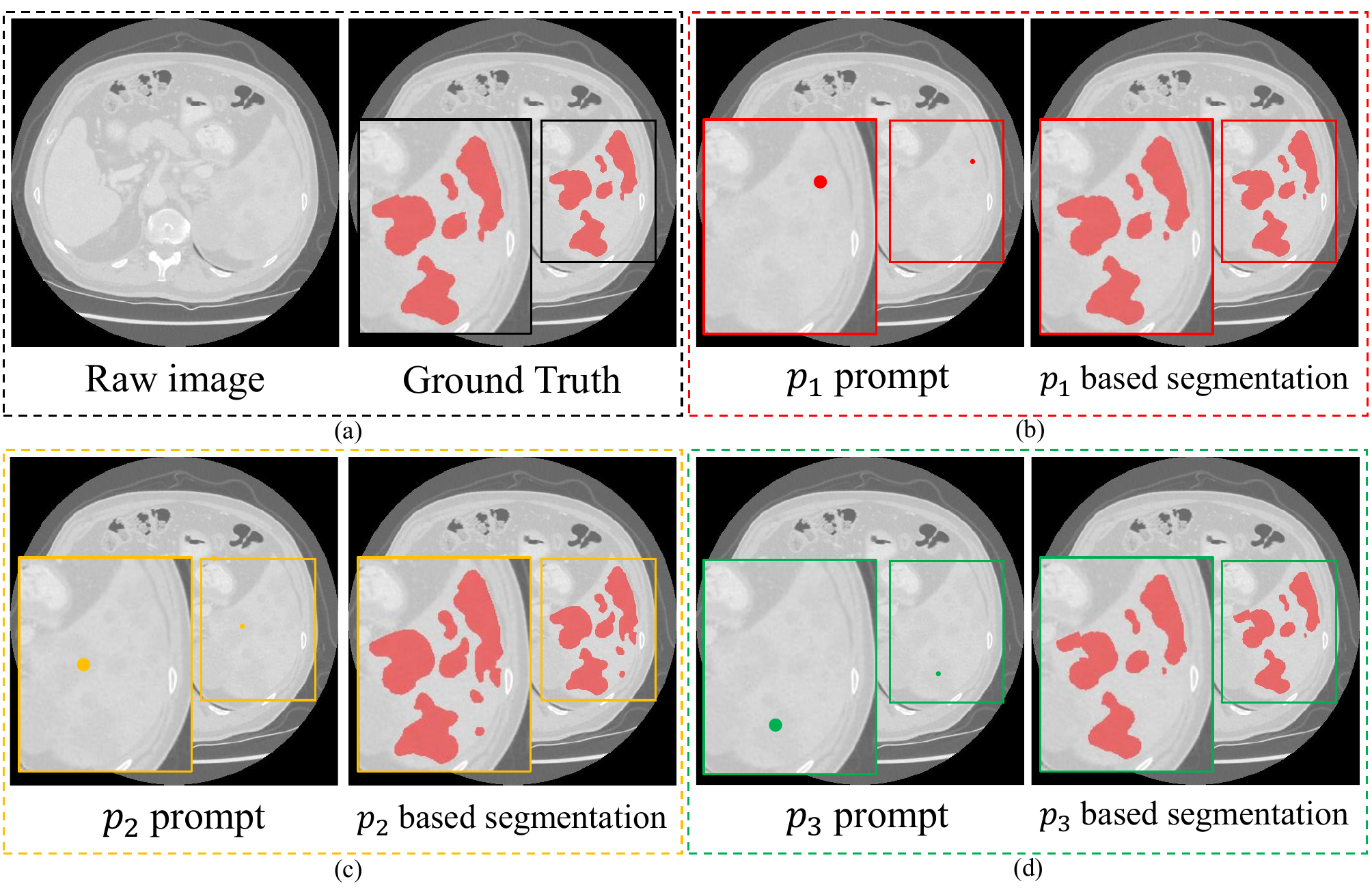}
\caption{Segmentation results under different single-point prompts} \label{fig:ablation}
\end{figure}

To comprehensively evaluate the effectiveness of each component within the proposed framework, we conducted a series of ablation experiments on the Liver Tumor dataset, as shown in Table~\ref{tab:ablation}. First, we evaluated the contribution of PC-IQG+CQRD. When these two modules were removed and the model reverted to the traditional single-instance segmentation paradigm, the Dice score dropped significantly from 60.47\% to 56.72\%. This 3.75 percentage point performance degradation resulted from the inability to simultaneously capture all similar instances in a scene during a single forward pass, leading to overall segmentation quality deterioration. Second, we validated the necessity of the hybrid encoder design by removing the CNN branch. Results showed that the model relying solely on the ViT backbone achieved a Dice coefficient of 58.95\%, a 1.52 percentage point decrease compared to the full model. More notably, the NSD metric plummeted from 74.61\% to 68.80\%, a significant drop of 5.81 percentage points. This outcome conclusively demonstrates the irreplaceable role of the CNN branch in capturing high-fidelity local details. Finally, we conducted a qualitative assessment of the framework's prompt robustness, as illustrated in Fig.~\ref{fig:ablation}. Within the same 3D image containing multiple liver tumors, we provided single-point prompts on different tumor instances (marked by differently colored dots in the figure) and observed the model's segmentation outputs. Experimental results demonstrate that regardless of which specific instance the user selects as the example, the model consistently identifies and segments all tumors of the same type within the scene. The predicted masks exhibit high consistency in both spatial distribution and the number of instances.

\section{Conclusion}
This paper proposes the MIQ-SAM3D multi-instance segmentation framework based on a hybrid CNN-Transformer architecture, achieving a breakthrough from single-point-to-single-mask to single-point-to-multi-instance segmentation. To the best of our knowledge, this is the first work applying prompt-conditioned instance query generation with competitive optimization decoders to 3D medical image segmentation. Our method designs a hybrid encoder that deeply integrates CNN-extracted boundary saliency features into ViT's self-attention computation via a spatial gating module, significantly enhancing the representation of fuzzy boundaries while preserving global semantic understanding. PC-IQG transforms user single-point prompts into dynamically generated object queries, while CQRD enables end-to-end parallel prediction of multiple instances through query-to-query competition. Extensive experiments on Liver Tumor and Kidney Tumor datasets demonstrate that the proposed method surpasses existing state-of-the-art approaches in segmentation accuracy while exhibiting outstanding prompt robustness, providing a practical solution for efficient annotation of clinically prevalent multiple lesions.
\section{ACKNOWLEDGMENTS}
No funding was received for conducting this study. The authors have no relevant financial or non-financial interests to disclose.

\bibliographystyle{IEEEbib}
\bibliography{refs}

\end{document}